\documentclass[11pt,a4paper]{article}
\usepackage{acl}
\usepackage{times}
\usepackage{latexsym}

\usepackage{microtype}

\usepackage[utf8]{inputenc} % allow utf-8 input
\usepackage[T1]{fontenc}    % use 8-bit T1 fonts
\usepackage{hyperref}       % hyperlinks
\usepackage{url}            % simple URL typesetting
\usepackage{booktabs}       % professional-quality tables
\usepackage{amsfonts}       % blackboard math symbols
\usepackage{nicefrac}       % compact symbols for 1/2, etc.
\usepackage{microtype}      % microtypography
\usepackage{lipsum}
\usepackage{graphicx}
\graphicspath{ {./images/} }

\usepackage{microtype}
\usepackage{times}
\usepackage{soul}
\usepackage{url}
\usepackage[utf8]{inputenc}

\usepackage{graphicx}
\usepackage{amsmath}
\usepackage{booktabs}
\usepackage{algorithm}
\usepackage{algorithmic}
\usepackage[shortlabels]{enumitem}
\usepackage[TABBOTCAP]{subfigure}
\usepackage{multirow}
\usepackage{diagbox}

\def\DM{{\mathcal D}}
\def\EM{{\mathcal E}}

\def\IM{{\mathcal I}}

\def\MM{{\mathcal M}}

\def\SM{{\mathcal S}}

\def\TM{{\mathcal T}}

\def\d{{\bf d}}

\def\e{{\bf e}}

\def\0{{\bf 0}}
\def\1{{\bf 1}}

\newcommand{\ModelName}{ZED}

\date{}
\title{Efficient Zero-shot Event Extraction with Context-Definition Alignment}

\author{Hongming Zhang, Wenlin Yao, Dong Yu\\
Tencent AI Lab, Bellevue, USA\\
\texttt{\{hongmzhang, wenlinyao, dyu\}@global.tencent.com}}

\begin{document}
\maketitle
\begin{abstract}

% Event detection is the task of finding 
Event extraction (EE) is the task of identifying interested event mentions from text.
Conventional efforts mainly focus on the supervised setting.
However, these supervised models cannot generalize to event types out of the pre-defined ontology.
To fill this gap, many efforts have been devoted to the zero-shot EE problem.
This paper follows the trend of modeling event-type semantics but moves one step further.
We argue that using the static embedding of the event type name might not be enough because a single word could be ambiguous, and we need a sentence to define the type semantics accurately.
To model the definition semantics, we use two separate transformer models to project the contextualized event mentions and corresponding definitions into the same embedding space and then minimize their embedding distance via contrastive learning.
On top of that, we also propose a warming phase to help the model learn the minor difference between similar definitions.
We name our approach Zero-shot Event extraction with Definition (\ModelName). 
Experiments on the MAVEN dataset show that our model significantly outperforms all previous zero-shot EE methods with fast inference speed due to the disjoint design.
Further experiments also show that \ModelName~can be easily applied to the few-shot setting when the annotation is available and consistently outperforms baseline supervised methods.

\end{abstract}

\section{Introduction}\label{sec:intro}

% Introduction to the event extraction task
Event extraction, the task of identifying event mentions from documents and classifying them into pre-defined event types, is a fundamental NLP problem~\cite{grishman2005nyu}.
As a centric information extraction task, event extraction is the foundation of a series of event-centric NLP applications~\cite{chen2021event} including event relation extraction~\cite{DBLP:conf/emnlp/WangCZR20}, event schema induction~\cite{DBLP:conf/emnlp/LiZLCJMCV20}, and missing event prediction~\cite{DBLP:conf/emnlp/ChaturvediPR17}.

\begin{figure}
    \centering
    \includegraphics[width=\linewidth]{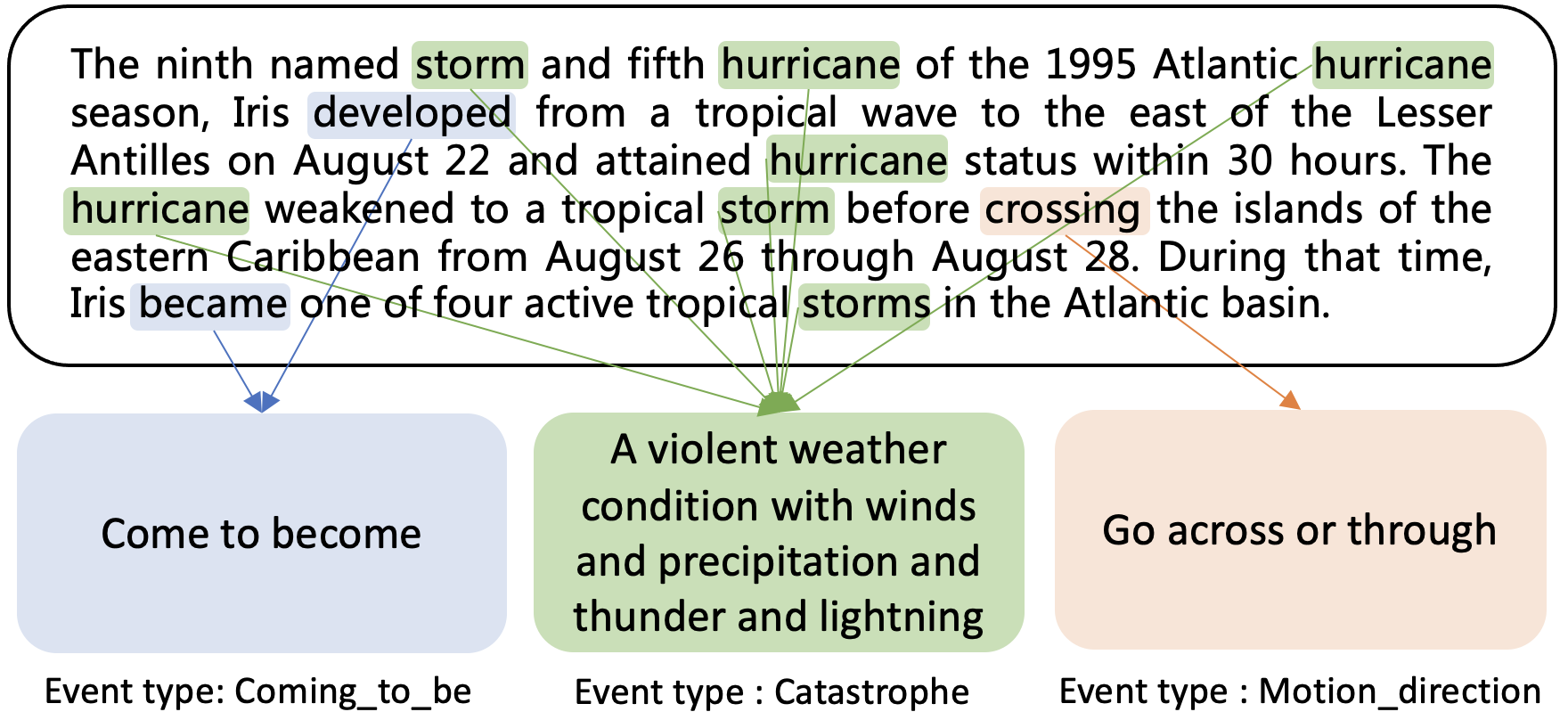}
    \caption{Zero-shot event extraction task demonstration. Given a corpus, the goal is to identify all event mentions that fit the target event definitions without using any annotation. The event definitions and corresponding mentions are indicated in the same color.}
    \label{fig:intro_demo}
\end{figure}

% Supervised event extraction systems and their limitations
Traditional event extraction efforts~\cite{DBLP:conf/emnlp/WaddenWLH19,DBLP:conf/naacl/WangHLSL19,DBLP:conf/acl/LinJHW20} mostly focus on learning to identify and classify events under a supervised learning setting, where a pre-defined event ontology and large-scale expert annotations is available.
However, the learned supervised models cannot be easily applied to new event types out of the pre-defined ontology, limiting these models' usage in real applications.

% Introduction to TE and QA models
Recently, large-scale pre-trained language models have demonstrated strong semantics representation capabilities and motivated a series of works to extract events in a zero-shot setting. For example, \citet{DBLP:conf/emnlp/DuC20} propose to manually design templates for each event type to convert the event extraction problem into a question-answering (QA) task and then leverage QA models to extract events. Following that, \citet{DBLP:conf/acl/LyuZSR20} propose to verbalize candidate triggers and event types into hypothesis and premises and leverage pre-trained textual entailment models to extract events.  
However, as analyzed in \cite{DBLP:conf/acl/LyuZSR20}, these models heavily rely on the template design and often suffer from the domain-shifting problem between the original training task and the new task. Moreover, as these models require jointly encoding the event mentions and event types, the time complexity is $O(N*T)$, where $N$ is the number of event mention candidates and $T$ is the number of event types.
Considering the low inference speed and high computation cost of inference with a deep model, such complexity could be a massive burden for real-time EE systems.

% Introduction to contextualized embedding
To avoid manually designing templates and to improve the inference efficiency, another line of work~\cite{DBLP:conf/acl/ZhangWR21} tries to leverage pre-trained language representation models (i.e., BERT~\cite{DBLP:conf/naacl/DevlinCLT19}) to acquire a contextualized event type representation. The model can decouple the mention and label representations during the inference time and predict the candidate trigger to the most similar event type based on the cosine similarity.
As a result, this method could significantly reduce the inference time complexity from $O(N*T)$ to $O(N+T)$. 
However, as the experiments show, using only the label name might not lead to a good event-type representation because the selected words could be ambiguous.
% As a result, the performance of this approach is limited.

In this work, we follow the trend of representation learning~\cite{DBLP:conf/acl/ZhangWR21,DBLP:conf/emnlp/GaoYC21} and move forward from representing each event type with a single name to a definition sentence. Specifically, we propose a three-stage event representation learning framework. In the offline pre-training phase, we leverage auto-extracted context-definition alignments to learn a definition encoding model that can encode the contextualized mentions and definitions into the same embedding space. In the second warming phase, we use the target event types to retrieve hard negative examples to further polish the model. In the end, we identify and classify event mentions based on the cosine similarity between the mention representation and corresponding event-type representations.
As our system is a disjoint model, the inference time complexity is also $O(N+T)$.
Experiments on MAVEN~\cite{DBLP:conf/emnlp/WangWHJHLLLLZ20}, the largest EE dataset to the best of our knowledge, show that \ModelName~outperforms all previous zero-shot approaches with high inference efficiency.
Further experiments show that \ModelName~could also be applied to the supervised setting, where it achieves comparable performance in the fully supervised setting and consistently outperforms baseline supervised models in the data-scarce learning settings.
Specifically, with 10\% of the training data, \ModelName~could achieve over 95\% of the full performance.
All the collected alignment data, created definitions, and the code are available at: https://github.com/tencent-ailab/ZED.

\begin{figure*}
    \centering
    \includegraphics[width=\linewidth]{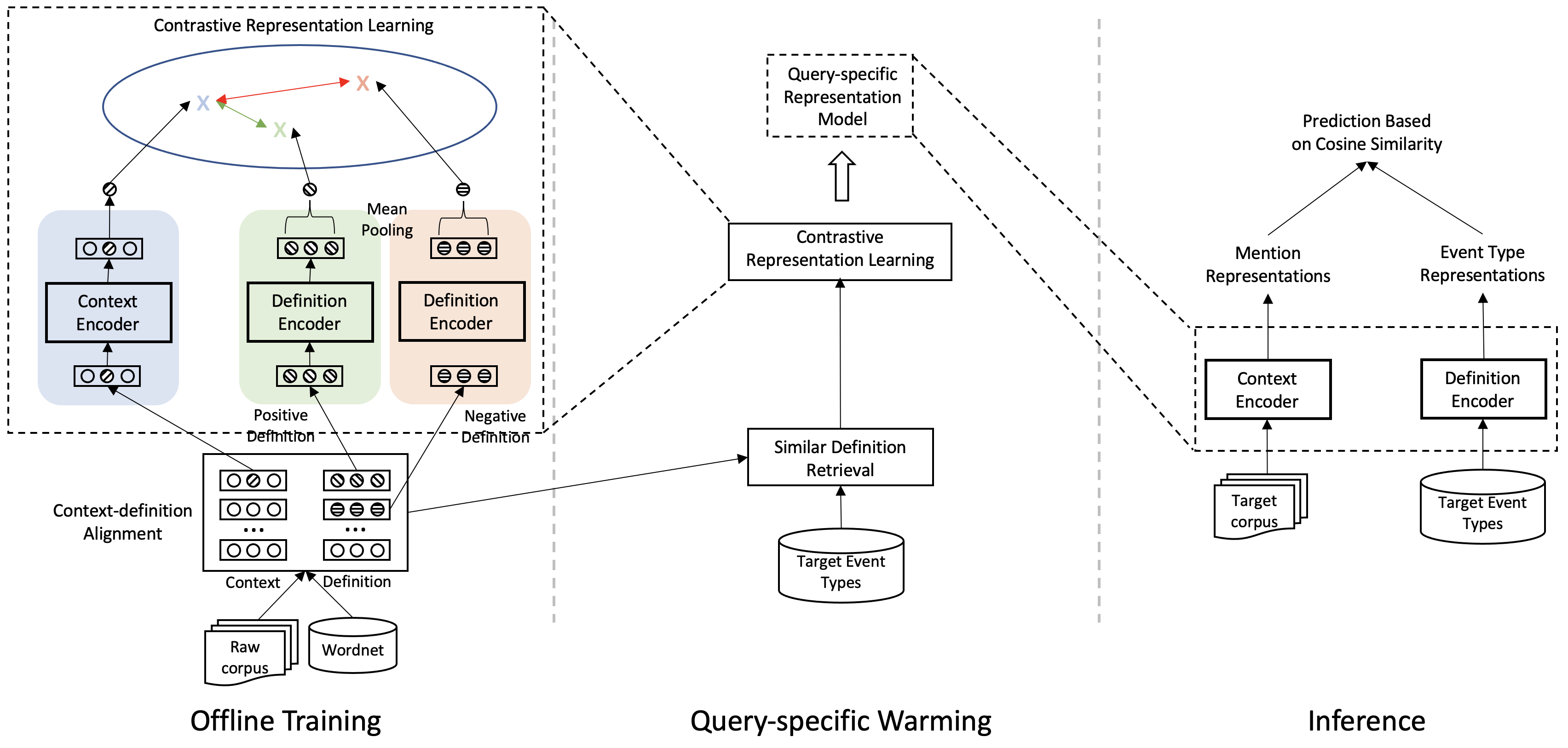}
    \caption{Overall framework of \ModelName. In the offline training phase, we train the separate context and definition encoders with auto-extracted context-definition alignment data. In the second warming phase, after knowing the target event types, as no annotation is provided, we first retrieve similar concepts from WordNet~\cite{miller1998wordnet} and use corresponding alignment data to polish the representation model. In the last inference phase, after encoding all candidate event type definitions, for each candidate event mention, we will encode it with the context encoder and determine whether it belongs to one of the target event types based on the cosine similarity.}
    \label{fig:framework}
\end{figure*}

\section{Related Works}

In this section, we introduce related works about event extractions, contrastive representation learning, and definition modeling.

\subsection{Event Extraction}

As a fundamental information extraction task~\cite{chen2021event}, event extraction has attracted many efforts in the NLP community~\cite{DBLP:conf/muc/Sundheim92,DBLP:conf/coling/GrishmanS96,DBLP:conf/aaai/Riloff96,grishman2005nyu,chen2021event,DBLP:journals/taslp/HongZYZ22}.
Recent success on the event extraction task mostly relies on employing either symbolic features~\cite{DBLP:conf/acl/JiG08,DBLP:conf/acl/LiaoG10,DBLP:conf/acl/LiuCHL016} or distributed features~\cite{DBLP:conf/acl/ChenXLZ015,DBLP:conf/naacl/NguyenCG16,DBLP:conf/emnlp/LiuLH18,DBLP:journals/dint/ZhangJS19,DBLP:conf/emnlp/WaddenWLH19,DBLP:conf/acl/LinJHW20} to learn supervised models with large-scale high-quality annotations.
However, the requirement of a pre-defined ontology and corresponding annotations limits the application of these models in real applications.

To address this issue and extract unseen event types, \citet{DBLP:conf/acl/DaganJVHCR18} propose a zero-shot event extraction task and use a transfer-learning framework to apply the model trained with seen event types to unseen ones.
However, the prerequisite of their high performance is the similarity between seen and unseen event types.
Recently, with the fast development of large-scale language models, several works~\cite{DBLP:conf/emnlp/DuC20,DBLP:conf/acl/LyuZSR20,DBLP:conf/acl/ZhangWR21} propose to leverage the pre-trained models to encode the label semantics either with templates or contextualized embeddings.
In this work, we follow the effort of using deep models to model the label semantics but make a step further.
Instead of directly using a pre-trained model, we train a disjoint context-to-definition alignment encoding model, which can effectively map the candidate event mentions and definitions into the same embedding space and thus more accurately and efficiently extract events for any arbitrarily defined event types.

% Unlike previous works, we do not use any annotation and only leverage the label semantics to classify event triggers and arguments.
% By combining our classification model and other NLP modules (i.e., SRL and mention detection), we achieve a decent zero-shot event extraction pipeline that can be easily applied to any new documents and event types.

\subsection{Contrastive Representation Learning}
% Introduction to our model

The contrastive loss~\cite{DBLP:conf/cvpr/ChopraHL05} is one of the most popular training objectives for representation learning. 
The original contrastive loss and its variations (e.g., triplet loss~\cite{DBLP:conf/cvpr/SchroffKP15}, lifted structured loss~\cite{oh2016deep}, N-pair loss~\cite{sohn2016improved}, and NCE loss~\cite{pmlr-v9-gutmann10a}) have been shown helpful for a series of vision applications~\cite{DBLP:conf/icml/RadfordKHRGASAM21}.
After being introduced to the NLP community, the contrastive learning-based method also leads to the success of a series of representation learning tasks such as sentence representation~\cite{DBLP:conf/emnlp/GaoYC21}.
Different from previous works, where the anchors and positive/negative examples typically belong to the same category (e.g., image/sentence), we propose to use the contextualized token representation as the anchor and event type definition representations as the positive/negative examples to better solve the zero-shot event extraction task.
Moreover, motivated by the success of the ``pre-training+fine-tuning'' paradigm, we propose a novel three-stage representation learning framework.

\subsection{Definition Modeling}

Humans are capable of understanding new concepts by reading their glosses or definitions. Thus, how to leverage the definitions and explanations from dictionaries to help understand human language is a long-standing question in the NLP community.
Most of the previous efforts in this direction are working on the word sense disambiguation task~\cite{DBLP:conf/acl/SuiCLXL18,DBLP:conf/emnlp/HuangSQH19,DBLP:conf/acl/BlevinsZ20,DBLP:conf/acl/KumarJST19,DBLP:conf/acl/BevilacquaN20,DBLP:conf/emnlp/YaoPJCYY21,su-etal-2022-multilingual,DBLP:conf/acl/SuZSZ22}.
These models learn to map a token into the correct pre-defined synset by either jointly or disjointly encoding the tokens and definitions.
Even though the setting of our model and these WSD models are similar, identifying event mentions that satisfy an arbitrary event type definition is a more challenging task~\cite{CoDA21}. WSD aims to learn to distinguish the correct synset versus several (typically less than 10) other pre-defined synsets, while our goal is to align an event mention and the corresponding definition, where all other arbitrary definitions are considered to be the negative candidates.
To address the engineering limitation that negative candidates exceed the GPU memory limitation, we propose a coarse-to-fine negative sampling strategy to help models learn the minor differences between similar definitions without forgetting the big picture.
% By doing so, we successfully train a descent context-definition alignment encoder.

% 

% Supervised text classification

% Previous zero-shot approaches

% This paper

\section{Task Definition}

We define the zero-shot event extraction task as follows. Given a document in the format of a sentence set $\SM$ and event type set $\EM$. Each event type $E \in \EM$ is defined with a natural sentence $d$. The task is to identify all mentions $\MM_E$ in $\SM$ that satisfy the definition of $E$ for each $E \in \EM$ without using direct annotations during the training phase.

\section{Model}

We present the model framework in Figure~\ref{fig:framework}. Motivated by the success of the ``pre-training + fine-tuning'' learning paradigm, we propose to address the zero-shot event extraction problem with a three-stage framework. Technical details are as follows.

\begin{table}[t]
\small
    \centering
    \begin{tabular}{p{3.2cm}|p{3.6cm}}
    \toprule
    Mention in Context     & Definition \\
         \midrule
        I \underline{love} playing basketball. & get pleasure from \\
         \hline
        Bob is \underline{studying} computer science. & be a student; follow a course of study; be enrolled at an institute of learning\\
         \hline
        I got \underline{promoted} after many years of hard work & give a promotion to or assign to a higher position\\
         \bottomrule
    \end{tabular}
    \caption{Demonstration of collected context and definition alignments. Target mentions are underlined.}
    \label{tab:event_description_demonstration}
\end{table}

\subsection{Offline Pre-training}

The offline pre-training step aims to train a decent definition encoder to map the target mention representation and corresponding definitions into the same embedding space. To achieve this goal, as no annotation is provided, we first collect context-definition alignments and then train the encoder with a contrastive learning loss.

\subsubsection{Data Preparation}

We select all verbal synsets from the WordNet ontology~\cite{miller1998wordnet} to form our open-world event definition set. In total, we collect 13,814 event synsets. After that, to collect large-scale alignment data between context and definitions, we apply the current state-of-the-art word sense disambiguation model~\cite{DBLP:conf/emnlp/YaoPJCYY21} to the NYT corpus~\cite{sandhaus2008new} to align tokens in NYT with their correct definitions.
We randomly select 10 context instances for each synset to speed up the training process.
As a result, we collect 775K context-definition alignments.
Examples of extracted alignments are presented in table~\ref{tab:event_description_demonstration}.

\subsubsection{Context-definition Alignment Encoding with Contrastive Learning}

The goal of the context-definition alignment encoding is encoding the contextualized representation of the target mention and the sentence representation of the definition into the same embedding space and pushing them to be closer to each other because they should have similar semantic meanings. 
As this objective aligns well with the learning objective of the contrastive learning framework, we follow the standard contrastive learning framework~\cite{DBLP:conf/cvpr/ChopraHL05}.
Specifically, we denote the pre-processed context-definition alignment set as $\TM$, where each instance $(S, i, j, D) \in \TM$ contains context sentence $S$, which is a list of tokens $w_1^S, w_2^S, ... , w_n^S$, target word starting position $i$, target word ending position\footnote{Each word could have multiple tokens because we follow the standard tokenization of BERT~\cite{DBLP:conf/naacl/DevlinCLT19}.} $j$, and a definition sentence $D$, which is also a list of tokens $w_1^D, w_2^D, ..., w_m^D$.
We follow the standard approach to get the contextualized word representation as the mean pooling of all sub-token representations:
\begin{equation}
    \e_{S,i,j} =\frac{\sum_{i \leq k \leq j}\e_k}{j-i+1}, 
\end{equation}
where $\e_k$ is the contextualized representation of token $k$ produced by a transformer baseline language model (e.g., BERT~\cite{DBLP:conf/naacl/DevlinCLT19}).
For the sentence encoding, we choose to use the average representation of all tokens as follows:
\begin{equation}\label{eq:definition_mean_pooling}
    \d_D = \frac{\sum_{1 \leq k \leq m}FFN(\e_k)}{m},
\end{equation}
where $FFN$ represents a two-layer feed-forward neural network and $\e_k$ is the token representation of token $w_k$.

Following the contrastive learning framework, during this step, we optimize the marginal ranking loss\footnote{We chose ranking loss over entropy loss mainly because the alignment data we used for the pre-training is automatically collected, and the training signal may contain noise. 
% A more detailed comparison between these two training objectives are presented in Section~\ref{sec:exp}.
}. Assume that the set of randomly sampled negative definitions is $\DM^\prime$, for each $D^\prime \in \DM^\prime$, we could follow equation~\ref{eq:definition_mean_pooling} to compute its representation as $\d_{D^{\prime}}$. For each instance $(S, i, j, D) \in \TM$ and a randomly sampled negative definition set $\DM^\prime$, we minimize the following marginal ranking loss:

\begin{equation}\label{eq:loss}
\small
    \frac{\sum_{D^\prime \in \DM^\prime} max (0, \epsilon -(cos(\e_{S,i,j}, \d_D)-cos(\e_{S,i,j}, \d_{D^{\prime}})))}{\|\DM^\prime\|},
\end{equation}
where $max$ means the maximum operation, $cos$ is the cosine similarity, and $\epsilon$ is the margin.

\subsection{Query-specific Warming}

After the pre-training phase, the model briefly understands how to project the contextualized event mentions and corresponding definitions into similar positions in the embeddings. However, its capability of distinguishing similar definitions is still limited because the previous negative sampling strategy does not encourage such capabilities. To address this issue, we introduce an additional warming phase to help models learn the minor difference between similar definitions.

Similar to how human beings understand new concepts by recalling relevant knowledge, we also retrieve relevant knowledge from $\TM$ to further fine-tune the model.
Specifically, assume that the set of interested event definitions is $\hat{\DM}$, for each $\hat{D} \in \hat{\DM}$, we first retrieve the most similar definition $\tilde{D}$ from the original definition set $\DM$ by:
\begin{equation}
    \tilde{D} = \arg \max_{D \in \DM} sim(PLM(D), PLM(\hat{D})),
\end{equation}
where $sim$ is the similarity measurement and $PLM$ represents the encoding with a pre-trained language model.
In our experiment, we select cosine similarity as the similarity measurement and average contextualized token embedding encoded with BERT-base~\cite{DBLP:conf/naacl/DevlinCLT19} as the encoding. But other techniques could also be applied.

We thus denote the set of all retrieved relevant definitions as $\tilde{\DM}$ and select a subset $\tilde{\IM}$ of $\IM$ such that all definitions in $\tilde{\IM}$ belong to $\tilde{\DM}$ to further fine-tune the model.
% To help the model learn to distinguish similar definitions, we introduce a mixed negative sampling strategy.
% Specifically, for each $\tilde{I} \in \tilde{\IM}$, we select a set of randomly chosen negative examples $\tilde{\DM}^\prime$ and a set of carefully selected strong negative examples $\tilde{\DM}^{\prime,s}$. To select $\tilde{\DM}^{\prime,s}$, we rank all $\tilde{D} \in \tilde{\DM}$ based on the cosine similarity between the mention representation and corresponding definition representation generated by the model: $cos(\e_{\tilde{S},\tilde{i},\tilde{j}}, \d_{\tilde{D}})$, where $\tilde{S}$, $\tilde{i}$, and $\tilde{j}$ are the context sentence, starting position, and end position in $\tilde{I}$.
% In the next step, we merge $\tilde{\DM}^\prime$ and $\tilde{\DM}^{\prime,s}$ to form the final negative example set.
% To help model the definition with a coarse-to-fine manner, we propose to gradually increase the probability of hard negative examples with the annealing strategy:
% \begin{equation}
% \small
%     p_{hard, k} = p_0 + (p_K-p_0) / (1 + \exp{(-(k-K/2))}),
% \end{equation}
% where $k$ and $K$ are the current fine-tuning iteration and total iteration numbers. $p_0$ and $p_K$ represent the initial and ending probability, respectively.
After generating all the data, we will fine-tune all models following the loss function in Equation~\ref{eq:loss}.

\subsection{Inference}

During the inference, we compute the representation for each candidate event mention in and target event type descriptions. After that, for each candidate mention, we compute its cosine similarity with all the target event-type representations. If the largest similarity is larger than a threshold $t$, this mention is identified and labeled as the most similar event type. Assume that the size of all candidate mentions and target event types are $N$ and $T$, respectively. Compared with previous zero-shot models that rely on the joint encoding of the candidate mention and target event types~\cite{DBLP:conf/emnlp/DuC20,DBLP:conf/acl/LyuZSR20,DBLP:conf/emnlp/YaoPJCYY21}, we successfully reduce the computation complexity from $O(N*T)$ to $O(N+T)$. A numerical evaluation of the computation efficiency is shown in Section~\ref{sec:efficiency}. 

\begin{table*}[t]
    \centering
    \small
    \begin{tabular}{l||ccc|ccc}
        \toprule
        \multirow{2}{*}{Model}  
                                &\multicolumn{3}{c|}{Identification} &\multicolumn{3}{c}{Identification+Classification} \\
                                & P & R & F1 & P & R & F1 \\
        \midrule
        Chance Performance & 19.33 & 20.05 & 19.68 & 0.11 & 0.11 & 0.11 \\
        Most Popular Event Type & 18.59 & 19.45 & 19.01 & 0.74 & 0.77 & 0.75 \\
        \midrule
        % Transfer Learning~\cite{DBLP:conf/acl/DaganJVHCR18}& & & & & & \\
        QA~\cite{DBLP:conf/emnlp/DuC20} & 19.76 & 45.18 & 27.49 & 4.19 & 9.58 & 5.83\\
        TE~\cite{DBLP:conf/acl/LyuZSR20} & 20.20 & 32.83 & 25.01 & 4.59 & 7.46 & 5.68 \\
        WSD~\cite{DBLP:conf/emnlp/YaoPJCYY21} & 24.66 & 80.52 & 37.76 & 5.36 & 17.51 & 8.21 \\
        CLE~\cite{DBLP:conf/acl/ZhangWR21} & 55.07 & 14.63 & 23.00 & 42.99 & 11.34 & 17.95 \\
        % Prompt (GPT-3) & & & & & & & & & & & & \\
        \midrule
        % Our work (two stage version) & 56.45 & 45.65 & 50.48 & 29.57 & 23.91 & 26.44 \\
        \ModelName & 59.37 & 42.28 & \textbf{49.39} & 39.63 & 28.22 & \textbf{32.96} \\
        \bottomrule
        \end{tabular}
    \caption{Zero-shot Event identification and classification results on MAVEN~\cite{DBLP:conf/emnlp/WangWHJHLLLLZ20}, which has 168 event types. Best F1 performances are indicated with bold font.}
    \label{tab:zero-shot_results}
\end{table*}

\section{Experiments}\label{sec:exp}

This section introduces experiment details, including the selected baseline methods, experiment datasets, and implementation details.

\subsection{Baseline Methods}

In the past two years, the community has been devoting significant effort to solving the zero-shot event extraction problem with different approaches. Specifically, we select the following best-performing models as our baselines.

\begin{enumerate}[leftmargin=*]
    \item \textbf{Pre-trained Question Answering Models}~\cite{DBLP:conf/emnlp/DuC20} (QA): Most NLP tasks can be converted into a QA format and event extraction is not an exception. Motivated by this, \citet{DBLP:conf/emnlp/DuC20} propose to design a question template for each target event type and directly ask a QA model to answer whether a mention is the target event.
    \item \textbf{Pre-trained Textual Entailment Models}~\cite{DBLP:conf/acl/LyuZSR20} (TE): Motivated by the QA approach, \citet{DBLP:conf/acl/LyuZSR20} explore the possibility of utilizing a pre-trained textual entailment (TE) model to automatically extract events. Specifically, for each target event type, \citet{DBLP:conf/acl/LyuZSR20} manually design a template to convert it into a hypothesis, treat the target event mention as the premise, and ask the TE model whether the target event mention can entail an event type.
    \item \textbf{Word Sense Disambiguation Models}~\cite{DBLP:conf/emnlp/YaoPJCYY21} (WSD): Prior WSD works also heavily rely on the correctly modeling of the definitions, so conceptually they could also be applied to the event extraction task following our setup. There are mainly two key differences between our work and ~\cite{DBLP:conf/emnlp/YaoPJCYY21}: (1) ~\citet{DBLP:conf/emnlp/YaoPJCYY21} encode the context and definition jointly while our model encodes them separately; (2) ~\citet{DBLP:conf/emnlp/YaoPJCYY21} aim at modeling the minor difference between different synsets of the same word while our work aims at modeling general definition semantics.
    \item \textbf{Contextualized Label Embedding}~\cite{DBLP:conf/acl/ZhangWR21} (CLE): The last baseline we compare with is the contextualized label representation. Specifically, for each target event type, ~\citet{DBLP:conf/acl/ZhangWR21} generate a contextualized label representation by putting the label name back into contexts and directly extracting events based on the similarity between the mention representation and event type representations. 
    % \item \textbf{Prompt with large Pre-trained Language models}~\cite{}:
\end{enumerate}

Besides these baselines, we also present the ``Chance'' performance, where a mention is randomly selected following the percentage of gold mentions and randomly assigned an event type, and the ``Most Popular Event Type'' performance, where a mention is also randomly selected following the percentage of gold mentions and is always predicted to be the most popular event type.

\subsection{Evaluation Dataset}

We select MAVEN~\cite{DBLP:conf/emnlp/WangWHJHLLLLZ20} as the evaluation dataset due to its large-scale and balanced distribution. Specifically, MAVEN contains 186 unique event types selected from FrameNet~\cite{baker1998berkeley} and 118,732 annotated event mentions, which is almost two magnitudes larger than the previous datasets such as ACE~\cite{grishman2005nyu}. Moreover, MAVEN provides the official event mention candidates to evaluate the mention understanding capability of all event extraction models more fairly.
As the original dataset only provides the event name in the format of a phrase (e.g., ``Body\_movement''), we directly use definitions from Wordnet as the description\footnote{For event names that have multiple synsets or not covered by WordNet, we manually select the most accurate description from WordNet.}. Examples of the event types and corresponding definitions are presented in Appendix Table~\ref{tab:maven_ontology}. 

\subsection{Implementation Details}

For baseline models, we conduct experiments with officially released code, hyperparameters, templates, and pre-trained models. For \ModelName, we use two separate encoders for the context and definition encoding. Both of them are initialized with BERT-base~\cite{DBLP:conf/naacl/DevlinCLT19}. As no training set is needed in the zero-shot setting and the test set of MAVEN is not publicly available, we report the performance on the dev set. Specifically, we set the margin to 0.2 for the marginal ranking loss and set the number of negative examples to 2. The selection threshold at the inference phrase is set to be 0.7. We train the model with ten epochs for both the pre-training and warming phrases. We directly evaluate the last checkpoint to simulate the real application, where no dev set is available. All models are trained with Tesla P40 with batch size 16. The pre-training and warming phrases will take around 200 and 3 hours on a single GPU, respectively, but we could speed it up with multiple GPUs. 

\begin{table}[t]
    \centering
    \footnotesize
    \begin{tabular}{l|c|c}
    \toprule
         & F1 (I) & F1 (I+C) \\
         \midrule
        \ModelName & 49.39 & 32.96 \\
        \midrule
        % $\,$ - Pre-training & & \\
        $\,$ - Warming & 47.86 (-1.53) & 21.89 (-11.07) \\
        $\,$ - Strong Negative Sampling                & 48.91 (-0.48) & 29.20 (-3.76) \\
    % \midrule
    % Ranking loss => Entropy loss & & \\
         \bottomrule
    \end{tabular}
    \caption{Ablation study. ``I'' and ``C'' represent the identification and classification, respectively.}
    \label{tab:ablation}
\end{table}

\section{Zero-shot Performance}

The zero-shot performance of all models is presented in Table~\ref{tab:zero-shot_results}, from which we can make the following observations:
\begin{enumerate}[leftmargin=*]
    \item All models significantly outperform the naive baselines even though they do not use any annotations. This observation shows that current deep models can indeed learn rich semantics that could generalize outside of their original training goal.
    \item The overall performance of pre-trained QA, TE, and WSD models is not satisfying because they suffer from domain shifting. For example, even though current deep-model-driven QA models have surpassed human performance on several leaderboards, they are still not ready to be used as a general QA model for solving tasks that require deep understanding, such as zero-shot event extraction.
    \item Compared with other methods, Contextualized label embedding achieves lower identification F1 but higher classification accuracy, which aligns with the original observation in~\cite{DBLP:conf/acl/ZhangWR21}. The reason behind this is that due to the cone property of the BERT representation (i.e., most of the token representations of BERT are grouped in a small region), it is tough to determine the cosine similarity boundary of whether an event mention fits a specific event type. As a result, even though CLE could accurately identify high-confident mentions, it cannot handle boundary ones very well.
    \item Compared with baseline methods, \ModelName~could perform better on both the identification and classification tasks. The main reason is that we are using definitions to model the label semantics, which is more accurate than a single word.
\end{enumerate}

\subsection{Ablation Study}

\begin{figure}[t]
    \centering
    \includegraphics[width=\linewidth]{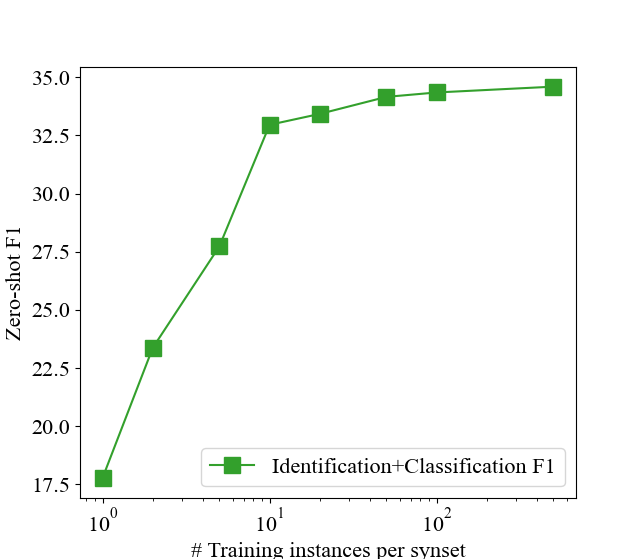}
    \caption{Effect of training instance number per definition. (Zero-shot performance on the Identification+Classification F1 is reported.)}
    \label{fig:training_instance}
\end{figure}

From the ablation study results in Table~\ref{tab:ablation}, we can see that if we remove the warming phase, the model's performance will drop on both the identification and classification, especially the classification step. This aligns well with our assumption that the model can learn to model the general definition semantics after the pre-training step but cannot distinguish minor differences very well. The performance drop of removing the strong negative sampling module indicates that strong negatives are crucial for the success of representation learning, which aligns well with previous observations~\cite{DBLP:conf/iclr/ClarkLLM20}.

Besides those ablation studies, we also show the impact of the pre-training data scale in Figure~\ref{fig:training_instance}. As expected, the more data we use, the better performance we will get. However, the performance gain after 10 instances per synset is limited. As a result, we select 10 instances for each synset as the pre-training data for training efficiency.

\subsection{Inference Efficiency}\label{sec:efficiency}

\begin{table*}[t]
    \centering
    \footnotesize
    \begin{tabular}{l||ccc|ccc}
        \toprule
        \multirow{2}{*}{Model}  
                                &\multicolumn{3}{c|}{Identification} &\multicolumn{3}{c}{Identification+Classification} \\
                                & P & R & F1 & P & R & F1 \\
        \midrule
        % Transfer Learning~\cite{DBLP:conf/acl/DaganJVHCR18}& & & & & & \\
        % DMCNN & & & & & & \\
        % Bi-LSTM + MLP & & & & & & \\
        % Bi-LSTM + CRF & & & & & & \\
        % MOGANED & & & & & & \\
        DMBERT&  73.37 & 87.82 & 79.95 & 61.20 &73.25 &66.69\\
        BERT+CRF & 75.19 & 81.80 & 78.35 & 64.40 & 70.28 & \textbf{67.21} \\
        % Prompt (GPT-3) & & & & & & & & & & & & \\
        \midrule
        % Our work (two stage version) & 56.45 & 45.65 & 50.48 & 29.57 & 23.91 & 26.44 \\
        \ModelName \ + Supervision & 82.48 & 80.76 & \textbf{81.61} & 67.87 & 66.46 & 67.16 \\
        \bottomrule
        \end{tabular}
    \caption{Model Performance with full annotations, Best F1 performances are indicated with the bold font. }
    \label{tab:supervised_performance}
\end{table*}

\begin{figure}
    \centering
    \includegraphics[width=\linewidth]{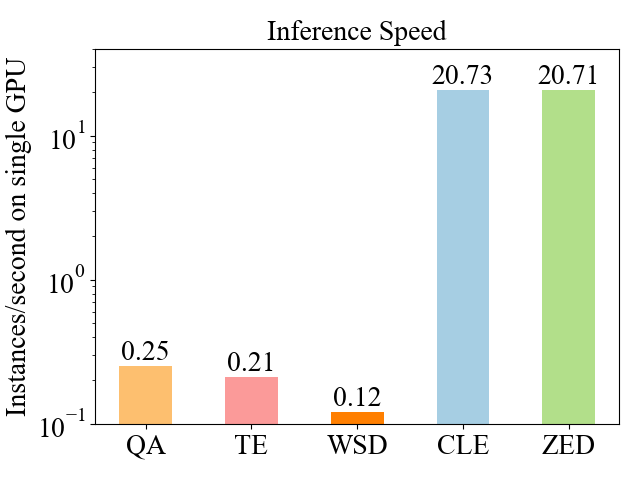}
    \caption{Inference Speed of all zero-shot models. For a fair comparison, we evaluate all models with the same GPU and use batch size 1. As our model is smaller than baseline models, we could use a larger batch size in real applications to further boost efficiency.}
    \label{fig:inference_speed}
\end{figure}

We present the inference speed of all evaluated models in Figure~\ref{fig:inference_speed}. As \ModelName~adopts a disjoint encoding design, we successfully reduce the computation complexity from $O(N*T)$ to $O(N+T)$, where $N$ is the number of event mentions and $T$ number of event types. On Maven, which has 168 different event types, \ModelName~could speed up the inference efficiency by almost two magnitudes.

% \subsection{Impact of Training Instance Number}

% \subsection{Precision versus Recall}

% \Red{As the threshold $\lambda$ plays an important role in the inference role, we plot the precision, recall, and F1 on the }

\section{Warming with Gold Annotation}
\ModelName~can also be adapted to a fully supervised or few-shot learning setting when the annotation is available. Specifically, during the warming phase of our model, we can replace the auto-retrieved examples with the annotated ones and fine-tune the model. In this section, we follow the benchmark paper~\cite{DBLP:conf/emnlp/WangWHJHLLLLZ20} to compare with the recent language model-driven baselines\footnote{Several other recent works (e.g., OneIE~\cite{DBLP:conf/acl/LinJHW20}) further improves the performance on event extraction by utilizing the constraints between trigger and arguments. However, as such information is not available in MAVEN and extracting arguments is beyond the research scope of this paper, we cannot compare with them.}: DMBERT~\cite{DBLP:conf/naacl/WangHLSL19} and BERT~\cite{DBLP:conf/naacl/DevlinCLT19} + CRF~\cite{DBLP:conf/icml/LaffertyMP01}, which also achieved the previous best performance. Please refer to the original papers for technical details of these baseline models.
We implement all models with the officially released code\footnote{https://github.com/THU-KEG/MAVEN-dataset} and report the average performance of five trials on the development set. All models are trained for ten epochs, and the final model is evaluated. Like the zero-shot setting, we also report the micro precision, recall, and F1 for both the ``identification'' and ``identification+classification'' settings. All hyper-parameters are based on the officially released code.

\begin{figure}
    \centering
    \includegraphics[width=\linewidth]{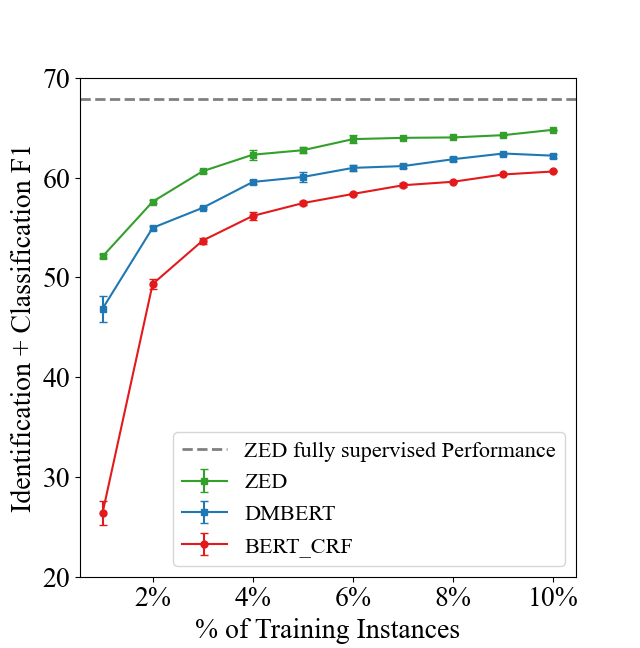}
    \caption{Model performance with limited annotation.}
    \label{fig:few-shot}
\end{figure}

Results in Table~\ref{tab:supervised_performance} show that with the help of the pre-training step, our model can outperform all previous supervised models on the identification task and comparable performance on the classification task. This makes sense because a carefully designed deep model could learn to identify and classify event mentions well with the large-scale annotation provided by MAVEN.

However, we argue that such a large-scale annotation is often expensive in terms of money and time. The data-scarce learning setting might be more applicable in real applications. Thus, we also test the performance of these supervised settings under the data-scarce learning setting.
Specifically, we randomly select 1\% to 10\% of the training sentences to be sampled from the training data and report the performances in Figure~\ref{fig:few-shot}. Our model can constantly outperform baseline models with a small number of annotations.
Especially when only 1\% of the data is available, we only have 7.07 mentions per event type, \ModelName~could achieve over 50 F1.   
With 10\% of the training data, \ModelName~could achieve over 95\% of full supervised performance.
These observations show that our framework could be applied to broader applications where limited or enough annotations are available besides the zero-shot setting.
Besides that, another interesting finding is that even though ``BERT+CRF'' could outperform ``DMBERT'' slightly when enough annotation is available, which is consistent with the observations in~\cite{DBLP:conf/emnlp/WangWHJHLLLLZ20}, its performance is worse under the data-scarce setting.
This observation indicates that using CRF might not be the optimal option when the annotation scale is limited.

% \subsection{Case Study}

\section{Conclusion}

This paper proposes a novel zero-shot event extraction framework \ModelName. Given a set of interested event types in the format of definitions, \ModelName~could automatically extract all the event mentions that fit the definitions from raw documents much better than previous methods.
Experiments show that the proposed warming phase and the mixed strong negative examples sampling strategies contribute to the success of \ModelName.
Additional experiments also show that \ModelName~could be applied to the supervised setting. Thanks to the pre-training phase, it could achieve good performance under both the fully supervised and data-scarce settings.

\section*{Acknowledgements}

We thank anonymous reviewers for their insightful comments and suggestions.

% \clearpage
\bibliographystyle{acl_natbib}
\bibliography{main}

\clearpage
\appendix

\section{MAVEN Ontology Demonstration}
\begin{table}[h]
    \centering
    \small
    \begin{tabular}{p{2cm}|p{4.7cm}}
    \toprule
       Name  & Definition \\
         \midrule
        Manufacturing & make or cause to be or to become\\
        Achieve & to gain with effort \\
        Communication & express in words \\
        Employment & engage or hire for work \\
        Process\_start & take the first step or steps in carrying out an action\\
        Theft & take without the owner's consent\\
        Legal\_rulings & pronounce a sentence in a court of law \\
        Influence & have an effect upon \\
        Give\_up & give up, such as power, as of monarchs and emperors, or duties and obligations\\
        Catastrophe & a violent weather condition \\
         \bottomrule
    \end{tabular}
    \caption{Representative MAVEN event types and associated definitions. All used definitions will be released with the code.}
    \label{tab:maven_ontology}
\end{table}
%\input{appendix.tex}

% \section{Effect of Pre-training Data Scale}\label{sec:data_scale}

% In this section

\end{document}